\documentclass[conference]{IEEEtran}
\IEEEoverridecommandlockouts
\usepackage{cite}
\usepackage{amsmath,amssymb,amsfonts}
\usepackage{algorithmic}
\usepackage{graphicx}
\usepackage{textcomp}
\usepackage{xcolor}
\newcommand{\changed}[1]{\textcolor{black}{#1}}
\def\BibTeX{{\rm B\kern-.05em{\sc i\kern-.025em b}\kern-.08em
    T\kern-.1667em\lower.7ex\hbox{E}\kern-.125emX}}
\begin{document}

\title{Similarity-based data mining for online domain adaptation of a sonar ATR system}
\author{\IEEEauthorblockN{Jean de Bodinat}
\IEEEauthorblockA{
\textit{Seebyte Ltd}\\
Edinburgh, Scotland \\
\small{jean.de-bodinat@seebyte.com}}
\and
\IEEEauthorblockN{Thomas Guerneve}
\IEEEauthorblockA{
\textit{Seebyte Ltd}\\
Edinburgh, Scotland \\
\small{thomas.guerneve@seebyte.com}}
\and
\IEEEauthorblockN{Jose Vazquez}
\IEEEauthorblockA{
\textit{Seebyte Ltd}\\
Edinburgh, Scotland \\
\small{jose.vazquez@seebyte.com}}
\and
\centering
\IEEEauthorblockN{Marija Jegorova}
\IEEEauthorblockA{
\textit{University of Edinburgh}\\
Edinburgh, Scotland \\
\small{m.jegorova@ed.ac.uk}}
}

\maketitle

\begin{abstract}
\changed{Due to the expensive nature of field data gathering, the lack of training data often limits the performance of Automatic Target Recognition (ATR) systems. This problem is often addressed with domain adaptation techniques, however the currently existing methods fail to satisfy the constraints of resource and time-limited underwater systems. We propose to address this issue via an online fine-tuning of the ATR algorithm using a novel data-selection method. Our proposed data-mining approach relies on visual similarity and outperforms the traditionally-employed hard-mining methods. We present a comparative performance analysis in a wide range of simulated environments and highlight the benefits of using our method for the rapid adaptation to previously-unseen environments.}
\end{abstract}


\section{Introduction}

Current state-of-the-art object detection and recognition methods rely on the offline training of deep learning models and require very large and generic datasets \cite{b1}. In order to deploy these in the real world, it is necessary to perform an offline training of these models with datasets that incorporate the specificities of the environments the systems are deployed in. When performing this offline training, a trade-off is typically made by choosing generic-enough data to suit as many types of input data as possible. This effectively achieves better genericity at the cost of performance on a particular domain \cite{b2}. On the other hand, the inevitably limited amount of data available at-hand limits the diversity of these training datasets. Offline training is therefore unavoidably sub-optimal.

In particular in the underwater domain, acquiring field data remains a time-consuming and expensive process due to the need of specialised equipment and trained operators. In addition to this, the validation and annotation of such datasets further limits the amount of useful data. As a result, only small and specific datasets are generally available to marine robotics engineers and researchers. Considering this very limited availability of the training data versus the vast variety of environments where autonomous underwater vehicles (AUVs) can be deployed, the real-world performance the ATR algorithms is often insufficient due to some of environments being under-represented in the initial training data. In particular, changes in the seafloor textures have been shown to have a significant impact on the performance of the ATR algorithms \cite{b3}. Variations  in  noise  levels  typical for  different  sensors  and  environments  tend  to  make  the predictions  about  the  ATR  performance  in  the  real-world scenarios  difficult. In addition to the unavoidable mismatch between training and operational data, generalisation often comes at the cost of lower accuracy on some of the domains \cite{b2}. These limitiations result in a need for online adaptation mechanisms to new sonar data, potentially issued from previously-unseen domains.

During a seafloor inspection detecting objects of interest, as well as conducting online imagery classification requires the constant attention of expert operators. In practise focus depletion inevitably leads to human mistakes. In this situation ATR algorithms can assist human operators with target labelling, effectively reducing stress and tiredness. Due to the variety of seabed textures that can be encountered and the limitations of offline training, the ability to adapt the response of ATR algorithms using live ground-truth labels from the operator is essential. We address this issue with an online method for fine-tuning an ATR algorithm using currently observed data. In this work we present a novel method to select the training data based on visual similarity with the currently observations. We benchmark our method against two standard data selection methods, commonly used to train deep learning models. We examine the strengths and weaknesses of the different data-mining strategies and demonstrate the benefits of using of using our similarity-based approach for rapid adaptation to previously-unseen environments.

\section{Background}
\label{sec:background}

While the applications of \textbf{domain-adaptation} have not been widely studied on sonar imagery, it has been a common topic for research in the fields computer vision and robotics \cite{b15, b14} and in particular in the autonomous driving car domain. Variations in city landscapes and weather conditions poses a challenge when training general-purpose models due to the lack of data for some of these (rare) domains and the predominantly offline training of these autonomous applications. 

Such data shortages are usually addressed by expanding on the existent small datasets with data augmentation techniques or domain translation (modifying the characteristics of data to match these of a different domain). Translating datasets has been done using \textbf{autoencoders} such as the MCAE \cite{b8} or \textbf{optimal transport} \cite{b9} and \textbf{style transfer methods} \cite{b11} such as Deep Photo \cite{b10}. It has also been achieved through the use of Generative Adversarial Networks (GAN) such as CycleGANs \cite{b6}, pix2pix \cite{b5} and AugCGan \cite{b7}. In particular, realistic underwater sonar imagery generation has been achieved by Markov-Conditional pix2pix \cite{MC-pix2pix} and R2D2-GAN for sonars of large resolution \cite{R2D2}. Most of these methods fall under the Unsupervised Domain Adaptation (UDA) category and work under the assumptions that we know beforehand what domain we will encounter. In addition to this, they are based on the assumption that large amounts of data from both domains are available and they typically require a large amount of time and computational resources for training. 

These assumptions do not necessarily hold in the context of seafloor inspection using sonar data as we cannot generally predict what type of seafloor that will be encountered during a mission. Similarly, due to the variability in hardware configurations as well as environmental variations, the amount and type of the noise in the measurements is expected to fluctuate. We therefore assume the possession of a limited amount of field data available for training before the mission. We also restrict the problem to online processing methods only where any processing must be performed under the time and resources constraints of a mission carried by an AUV. 

Machine learning approaches being data-driven, the availability of data availability is a common problem  and has been approached in different ways. \textbf{Model-driven} methods have been developed in order to take advantage of the fact the small amount of available labelled data from the target domain by designing neural networks composed of a domain-invariant feature extractor part and a small domain-specific tunable component that typically require small amounts of data. These \textbf{Few-shot learning} methods \cite{b12}, for instance Few-Shot Adaptive Faster R-CNN \cite{b17} are fast, domain-adaptive and stable. They are however offline methods and based on out-performed object detection architecture such as the R-CNN \cite{b16}. Current state-of-the-art object detection methods such as YOLO \cite{b18} and retinanet \cite{retina} architectures have focused on increasing accuracy and training speed rather than flexibility and adaptability to rare domains \cite{b17}. They however greatly out-perform previous state-of-the-art methods like the R-CNN on representative datasets \cite{b18}. 

While addressing domain adaptation, these state-of-the-art methods remain ill-suited to the following conditions required for the successful operation of the underwater sonar ATR systems:
\begin{itemize}
    \item \textbf{Scarce data:} The amount of available data for pre-mission training is very limited (typically a few hundreds of images with a limited range of objects of interests). In addition to this pre-training dataset, we aim for taking advantage of data acquired during the mission in an online fashion. In this paper, we consider missions as containing 50 to 200 images of 1000x500 pixels, with a 20\% chance of an image containing an object of interest. This poses a significant challenge for \textbf{adversarial techniques} that usually require large amounts of target domain data.
    \item \textbf{Time:} The amount of time required by an experienced human operator to label one of the image is estimated to 10 seconds. Therefore processing a mission by an operator would take between 8 and 35 minutes. A performance improvement would need to comply with these time-constraints and the limited resources of the autonomous vehicle, typically hardware platforms of a previous generation. Most of the style transfer methods are time-consuming to train, making these inadapted to our use-case.
    \item \textbf{Prior knowledge on the domain:} no prior information about the operation environment is available. Moreover, domains (i.e., terrain, water temperature and currents, as well as noise sources) can abruptly change from one image to another. State-of-the-art domain adaption techniques operate in an offline manner thus requiring at least some prior knowledge of the domain to adapt to.
    \item \textbf{Online Supervised learning: } human operator labelling a current mission provides us with some on-the-fly labels as the mission progresses. This can be used to our advantage improving the mission-specific ATR performance within the course of a mission. Most style transfer methods are not designed to use such on-the-fly information at all.
 \end{itemize}

In order to increase performance of our ATR model on new domains, we compromise between a model-driven approach and data-driven approach. We optimize the training procedure of a classical object detector neural network and adapt it to online training so as to take advantage of the latest labeled data. This makes possible the convergence towards a domain-dependent optimum during the mission at the expense of a slight loss in genericity. Indeed, insteaf of generating data for a new domain and training a general-purpose model, we follow the principle of the "no-free lunch" theorem \cite{b2} and train a model to be specialized and fine-tuned to the the current mission domain. Our work focused in particular in optimising the online training data selection by investigating the use of commonly employed methods such as hard-mining techniques \cite{b19} and developing our own selection method based on visual similarity. We also augment this training data using fast non-deep learning, making the processing feasible on a embedded platform in real time conditions.

\section{Method}
\label{sec:method}

\subsection{Online Learning Framework}
\label{sec:olframework}
\begin{figure}[htp]
\centering
\includegraphics[width=1\linewidth]{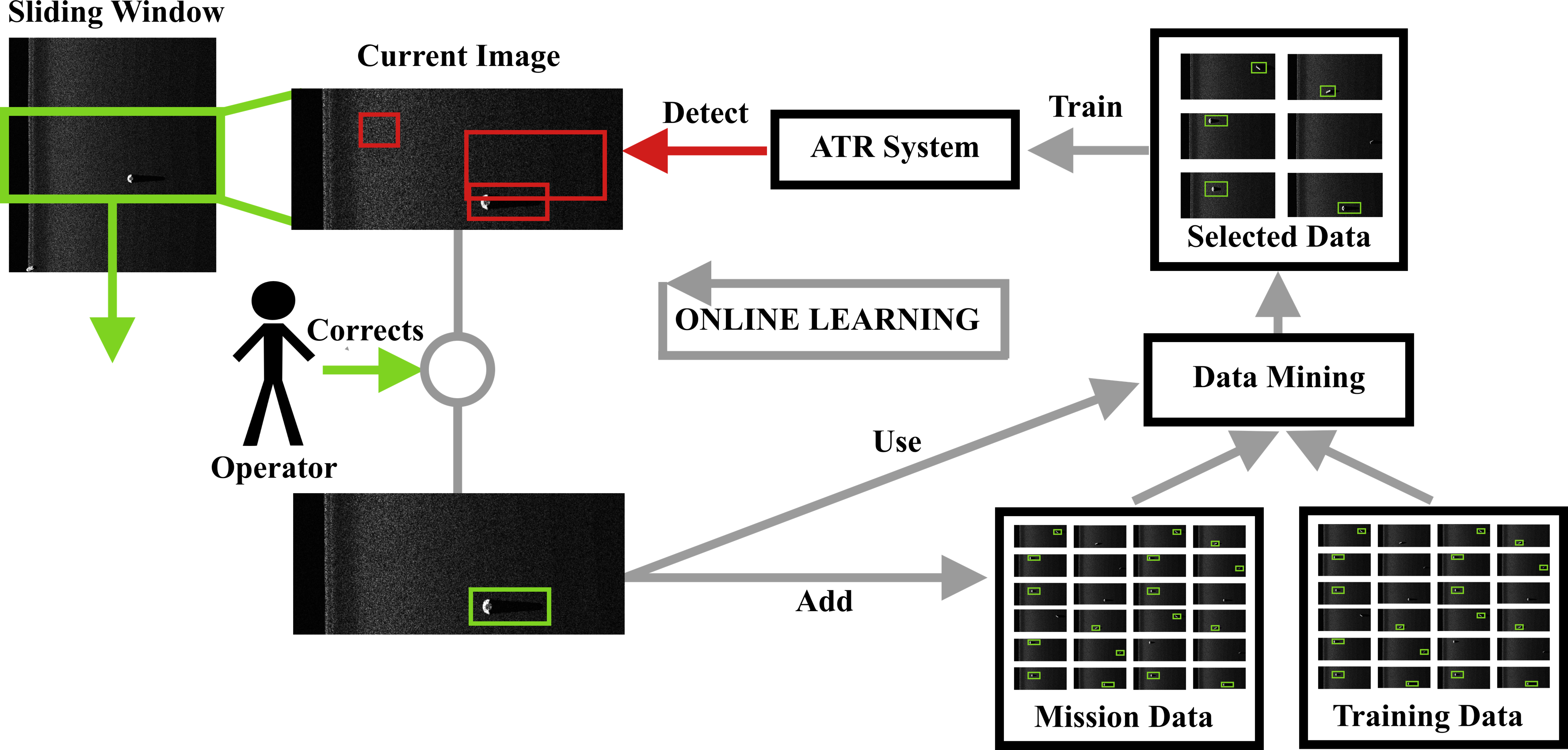}
    \caption{\textbf{Overall online learning framework.} As the sonar data gets checked for the presence of objects of interest both ATR and the human operator locate and label these objects. The human-operator labels are providing the new ground truth, used to correct the ATR labels. Human-labelled data gets added to the general pool of the training data and then selected by data-mining algorithms to be used for the online fine-tuning the ATR system. The update cycle keeps running throughout the entire mission, resulting in a constant domain adaptation throughout the mission.}
\label{fig:ol}
\end{figure}

Figure~\ref{fig:ol} provides a high-level illustration of the learning framework we adopt. The core of the method is an online learning framework aimed at converging to a domain-specific optimal weights (for the ATR neural network) within a mission run-time. This framework uses live labelling from a human operator at each mission frame to train the network with this new ground-truth as well as a carefully selected set of already-labelled frames, coming from the available training set. Additionally, it allows for using a variety of data mining techniques to select this set, on both the mission dataset (labelled on-the-fly) and the training dataset in a plug-in, plug-out fashion (see Figure \ref{fig:ol}). Multiple combinations of data mining methods on both datasets are therefore possible. In this work we study the impact they have on overall performance during a mission with the aim of improving the mAP (Mean Average Precision) metric, a commonly employed metric in machine learning.

In order to augment the available data, we introduced a few fast and commonly-employed techniques such as geometric transforms or blurring. These can be executed on-the-fly without prior-training and allow us to artifically increase the diversity of the dataset.
The crucial component of our method is the similarity score that enables us to select a set of frames based on visual similarity to the latest seen frame. Therefore throughout regular retraining, the current observations enable us to refine our model and converge to a domain-dependent optimum by dynamically adapting to the latest seen domain. The similarity score therefore encapsulates the domain description and determines the next state of the model.

\subsection{Similarity Metric}
\label{sec:simmetric}
\begin{figure}[htp]
\centering
\includegraphics[width=1\linewidth]{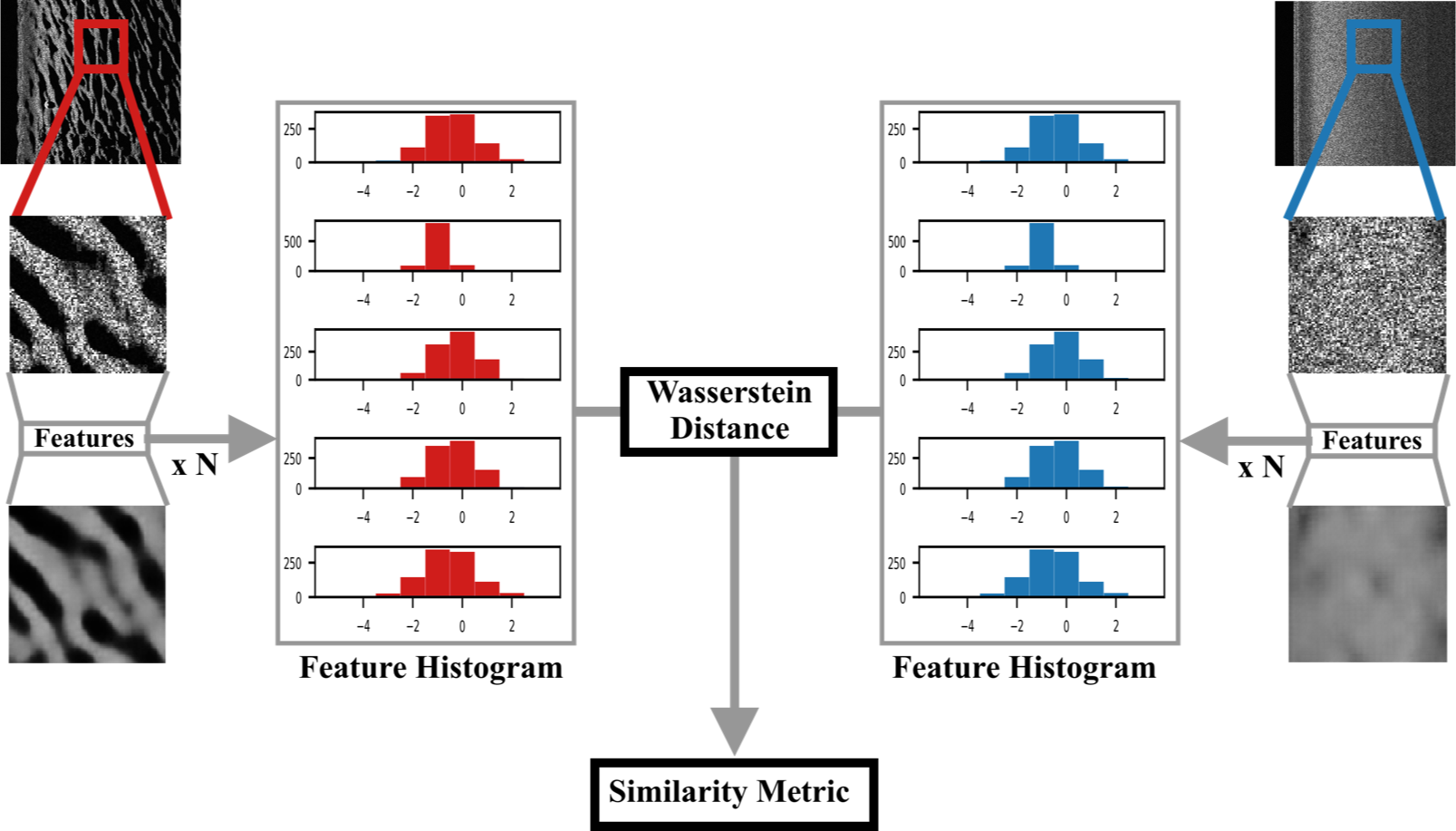}
    \caption{\textbf{EMD-based similarity metric explained.} First, $128\times128$ snippets get sampled from the currently inspected sonar image. Then they get converted into features via an autoencoder. Finally, the similarity of feature distributions between the currently-inspected image (current domain) and the previously available data (other previously observed domains) gets assessed and represented by an Earth Mover's Distance (EMD) akin to a Wasserstein distance. The most similar images are then selected to fine-tune the ATR system for the current domain.}
\label{fig:similarity}
\end{figure}
The seafloors can be accurately classified using textures in sonar data images. With that in mind, we have developed a similarity metric by extracting textural features from sonar image. Our algorithm, illustrated in Figure \ref{fig:similarity}, starts by cropping $N$ snippets from an image, each image being $500\times1000$ pixels, and snippets being $128\times128$ pixels. The dimensionality of these snippets then gets reduced by an autoencoder (pre-trained with COCO dataset \cite{b12} converted into black-and-white images). As a result, we get $M$ feature vectors. We then bin these feature vectors and build a distribution vector for each image. This gives us an array of probability distribution over the textural characteristics of our image that we use to characterise each sonar image. Given two distribution arrays evaluated on two sonar images, we compare each element of the array using the Earth Mover's Distance \cite{b13} which computes the Wasserstein distance between two probability distributions. This results in a 64-sized vector of distances of which we then take the L1 norm to compute our final textural feature distance metric. 


\subsection{Data-mining methods}
\label{sec:datamining}

In this section we describe the data selection approach implemented in our work:
\begin{itemize}
\item Sliding window: selecting the 10 last frames of the mission, effectively assuming slow variations in the environment.
\item Hard mining: selecting frames from the mission by using their latest recorded classification loss (Chapter \ref{sec:losssampling}) on the basis on prioritizing high loss values.
\item Similarity-based sampling: selecting frames by employing the similarity metric described in \ref{sec:simmetric} and considering the most similar images in our database (Chapter \ref{sec:similaritybased}).
\end{itemize}
The selected frames are a combination in equal proportions of frames issues from the pre-training dataset as well as recently-acquired images during the mission.

\subsubsection{hard-mining with live updates}
\label{sec:losssampling}
The hard-mining data selection method uses the latest recorded loss of the model on each frame of the mission, according to the following equation: 
\begin{equation}
\begin{array}{r c l}
    p(x=i|L)  & = & (L_{i}+\epsilon . \bar L)/(\sum_{j=0}^n(L_{j}+\epsilon . \bar L))  \\
\end{array}
\label{eq:losssampling}
\end{equation}

Where $L$ is the loss, $i$ and $j$ are indexes frames of the mission, $\bar L$ the average of all the frames losses and $\epsilon$ a variable which compensates for the difference between the loss of each frame. The higher $\epsilon$ is, the higher the chance to select frames with low losses is (exploration vs exploitation). This equation shows the probability of selecting a frame of index $i$ given its loss $L_{i}$. Using this equation and a new frame selection at each online iteration, we are able to dynamically select frames that are difficult for our model. 

\subsubsection{Similarity-based}
\label{sec:similaritybased}
The second method simply uses the similarity metric (Chapter \ref{sec:simmetric}) to select an $N$ number of frames, that are the most similar to the latest observed frame of the mission.

\section{Experimental Results}
\label{sec:xpres}

\subsection{Problem setup}
\label{sec:world}

\begin{figure}[htp]
\centering
\includegraphics[width=1\linewidth]{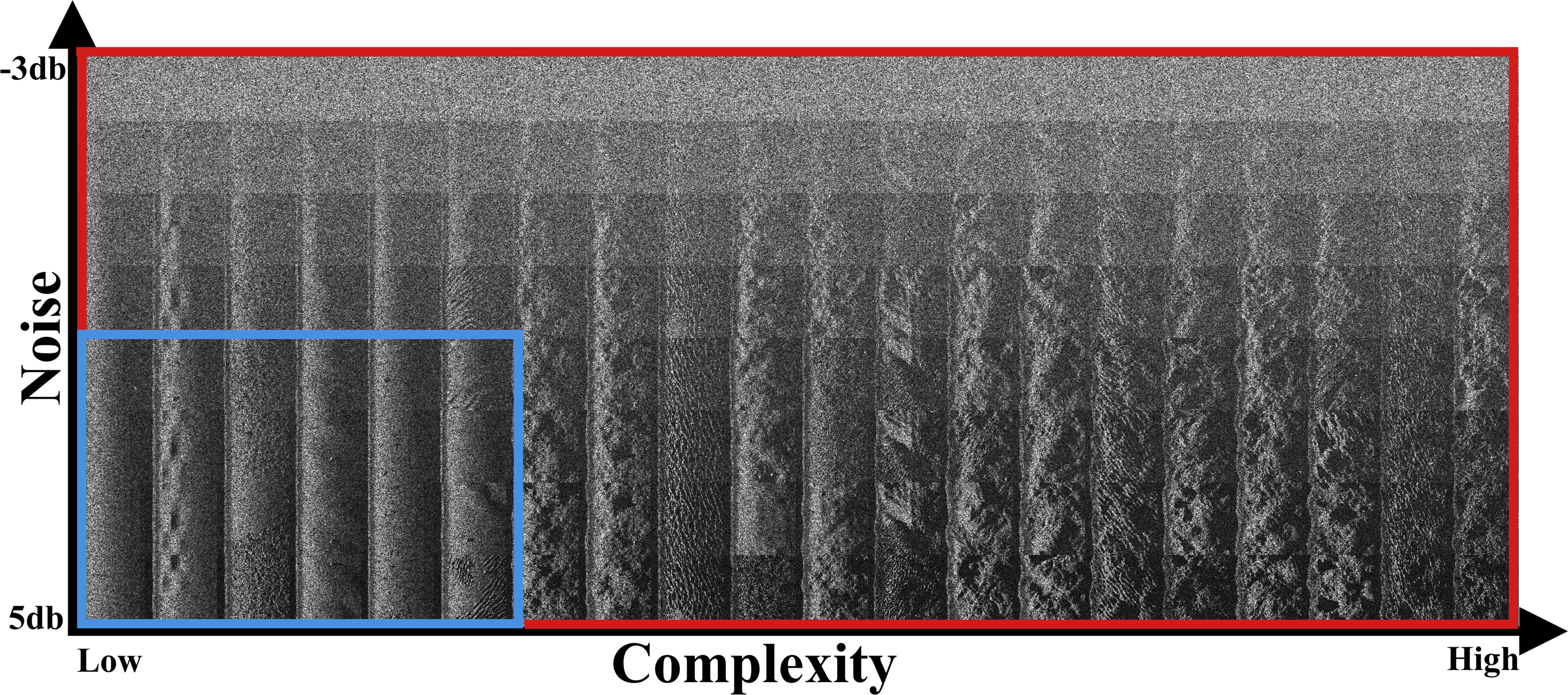}
    \caption{The 160 data domains used in the experiments. The vertical axis represents variations of the SNR while the horizontal axis represents the complexity of the seafloor. The blue subset represent the pre-training (offline training) dataset while the red set represents all the domains used in test missions.}
\label{fig:terrains}
\end{figure}

In order to test whether our assumptions are valid, we conducted some experiments on a purposely-designed simulated dataset. In order to generate data, a simple raytracing-based sidescan SONAR simulation tool with the ability to generate terrains of different types was employed. Our test was then generated by creating 160 different domains. Within each domain, 500 sidescan SONAR images were generated with a size of 1000x500 pixels and a square resolution of 5cm. In order to study the adaptibility to environment changes, the domains were generated with 8 different levels of sensor noise and with various types of seafloors and materials. As visible in Figure \ref{fig:terrains}, these domains can be reorganised as a 2D matrix were the dimensions are respectively the noise level and the complexity. While the noise level is a simple parameter of the simulation, the complexity of the data was empirically measured for each domain by training and testing on the same domain and evaluating the in-situ performance. The lower the classification performance, the more complex the domain was deemed to be. As a result, we obtain a 2D set of domains (red rectangle in Figure \ref{fig:terrains}) ranging from a simple (bottom left domain) to harder scenarios as we move up and right.

\begin{figure}[htp]
\centering
\includegraphics[width=1.0\linewidth]{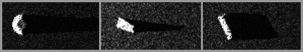}
    \caption{\textbf{Examples of simulated objects of interest.} These sidescan sonar views are characterized by an object shape (highlight) followed by shadow, both are used for identification by the ATR. Left: a cone; Middle: a wedge; Right: a cylinder.}
\label{fig:objects}
\end{figure}

In order to study the classification performance of our ATR, we inserted in these domains three typical types of objects: cones, cylinders and wedges as illustrated on Figure \ref{fig:objects}. Since objects are generally a rare occurrence in underwater sonar data analysis, we only inserted objects on 20\% of the dataset frames. 

Using this dataset we pre-trained our model with a subset of 49 low-complexity domains (represented by the blue rectangle on Figure \ref{fig:terrains}). These simple domains (mostly flat and low-frequency ripply terrain) were chosen for training as they are representative of the small amount of real data we possess. Finally, we generated a validation set composed of randomly generated missions of 100 to 200 frames, using one or multiple sequences of frames from one or multiple domains. 

The domain adaptation is then performed by retraining the ATR model every 10 frames on a custom set of images, composed of images coming from the pre-training dataset (blue rectangle) as well as images previously seen in the test mission (in equal proportions 10 frames from each dataset). Three methods of selecting data from the pre-training dataset were evaluated:
\begin{itemize}
    \item \textbf{None:} no data was taken out of the initial dataset.
    \item \textbf{Random:} randomly-selected images.
    \item \textbf{Similarity: } the most similar images were selected.
 \end{itemize}
Likewise, three methods of selecting data from the current test mission were evaluated: sliding window, loss-sampling and similarity-based as described in section \ref{sec:datamining}. In total, we therefore tested 9 data-selection approaches.

We then study the adaptability to new domains in two types of test scenarios: single-domain scenarios where we the test dataset is sampled from a single domain only and multi-domain scenarios where each test mission is composed of images issued from different types of domains. Finally we evaluate this adaptability by measuring the mAP (mean average precision) of our ATR over each test mission (100 frames per mission).
\subsection{Single-domain runs}
\label{sec:singledomain}
In this section, we present results obtained on single-domain test scenarios with the different data-mining approaches. As visible in Figure \ref{tab:singledomainresults}, the initial mAP is the same for each test case, we therefore present the variation in mAP occuring while retraining during the mission. 

\begin{table}[tp]
\begin{center}
\begin{small}
\begin{tabular}{|c|c|c|c|c|}
\hline
Data-mining & Data-mining & Initial & mAP & mAP \% \\
method on & method on & mAP & increase & increase\\
pre-training & test mission &  &  &  \\
dataset & data & & & \\
\hline

None & Sliding window & 81.3 & 2.2 & \changed{2.7} \% \\
Random & Sliding window & 81.3 & 5.3 & \changed{6.5} \% \\
Similarity & Sliding window & 81.3 & \textbf{6.9} & \changed{\textbf{8.5}} \% \\

None & Hard mining & 81.3 & 5.5 & \changed{6.8} \% \\
Random & Hard mining & 81.3 & 6.3 & \changed{7.7} \% \\
Similarity & Hard mining & 81.3 & \textbf{7.1} & \changed{\textbf{8.7}} \% \\

None & Similarity-based & 81.3 & 3.6 & \changed{4.4} \% \\
Random & Similarity-based & 81.3 & 5.8 & \changed{7.1} \% \\
Similarity & Similarity-based & 81.3 & 6.1 & \changed{7.5} \% \\

\hline
\end{tabular}
\end{small}
\vspace{5pt}
\caption{Classification performance on single-domain test scenarios}
\label{tab:singledomainresults}
\end{center}
\vspace{-25pt}
\end{table}

Table \ref{tab:singledomainresults} shows that all of the presented combinations of the data-mining techniques provide some improvement in the ATR system performance. As can be seen in the table, selecting data based on similarity provides the best results when selecting data from the pre-training data. On the opposite the best data selection method from the test mission data is the hard mining method.

\subsection{Multiple-domain runs}
\label{sec:multipledomain}

Since in the real world, seafloor types can change multiple times throughout a mission, we study here the impact on classification performance by running tests on missions containing data from different domains.

As can be seen in \ref{tab:multidomainresults}, the level of impact on performance on multi-domain scenarios is comparable to the single-domain scenario. Similarly to the single-domain scenario, similarity-based selection method provides the best mAP increase when selecting data from the pre-training dataset. Unlike the single-domain scenario, the similarity data-mining also provides the best results when used as a data-mining method on the mission data.
\begin{table}[htp]
\begin{center}
\begin{small}
\begin{tabular}{|c|c|c|c|c|}
\hline
Data-mining & Data-mining & Initial & mAP & mAP \% \\
method on & method on & mAP & increase & increase\\
pre-training & test mission &  &  &  \\
dataset & data & & & \\
\hline

None & Sliding window & 80.5 & -4.00 & \changed{-4.95} \% \\
Random & Sliding window & 80.5 & 3.60 & \changed{4.47} \% \\
Similarity & Sliding window & 80.5 & 3.80 & \changed{4.72} \% \\

None & Hard mining & 80.5 & 0.42 & \changed{0.52} \% \\
Random & Hard mining & 80.5 & 4.30 & \changed{5.34} \% \\
Similarity & Hard mining & 80.5 & \textbf{4.90} & \changed{\textbf{6.09}} \% \\

None & Similarity-based & 80.5 & 1.13 & \changed{1.40} \% \\
Random & Similarity-based & 80.5 & 4.80 & \changed{5.96} \% \\
Similarity & Similarity-based & 80.5 & \textbf{5.30} & \changed{\textbf{6.58}} \% \\

\hline
\end{tabular}
\end{small}
\vspace{5pt}
\caption{Classification performance on multi-domain test scenarios}
\label{tab:multidomainresults}
\end{center}
\vspace{-25pt}
\end{table}

\section{Discussion}
\label{sec:xpres}

Our experimental results show that the similarity-based method when used on the initial pre-training dataset is consistently increasing performance in all type of missions. When being employed at selecting data from the current mission, differences appeared between the single-domain and the multi-domain test scenario. When considering multi-domain missions, the mission data is intrinsically varied and therefore requires careful data-selection to match the currently observed data. On the opposite when considering single-domain scenarios, the textural similarity can be deemed to be equivalent across all images of the current mission (similar type of terrain across the mission) but the value of the samples in terms of training can be better estimated by the loss, resulting in better performance with the loss-mining approach. Similarity-based selection therefore appears to be well suited when selecting samples from a diverse dataset while loss-sampling is an objective sampling approach that remains strong for selecting data out of an homogeneous dataset (such as in the single-domain scenario).
This result also highlights the importance of specialised fine-tuning depending on the estimated knowledge gap. In a multi-domain context, the images previously seen during the mission might not describe well enough the currently observed terrain. In this situation and in particular with the goal of fine-tuning to the current domain, there is a need to compare the current observations and match them to previously-seen data in a loss-agnostic way since any out-of-domain image would give a high loss while being ill-suited to describing the current observations. On the other hand the results in single-domain scenarios show that for a given level of similarity (choosing samples out a set of similar images), the loss-based sampling remains a good data-mining method since in this context where the domain is well known (through previous observations) the loss characterises well the knowledge gap. These results therefore indicate that a two-step data-mining approach would be beneficial to both scenarios: a data subset could first be selected out of the complete dataset (pre-training and current mission) on the basis of visual resemblance using the similarity metric and then be refined using a loss-sampling approach. 
As could be expected, these tests show that performing an online fine tuning of the ATR consistently increases the performance over an offline training approach. While these results were observed with all data-mining methods, our work shows that adopting a similarity-based selection of the data provides an additional benefit of about 1\% on average over the commonly employed loss-sampling approach. 
While the amount of pre-training data is limited, we observed that its presence provides stability during the re-tuning by not only relying on the few labelled images available during the mission. In order for any data-mining method to be effective, a sufficient amount of data to compare to must be available. In our test scenarios, the pre-training dataset we used represented 15\% of the total number of domains used for testing and was representative of the domain distribution encountered in real datasets. This shows that this approach can be considered for embedded processing where only a relatively small amount of data can be kept on board. Due to the potentially high-variability of environments encountered during field missions and the impossibility of acquiring sufficiently diverse data for pre-training, we argue that this approach would translate well to real-case scenarios. Conversely, increasing the amount of on-board data available for mining should improve the performance on more complex domains.
\section{Conclusion}
\label{sec:concl}
In this work we have presented a dynamic domain adaptation method based on visual similarity. Our method is an improvement over the state-of-the-art hard-mining approach and benefits from efficient online selection of relevant training data. Our experiments showed that this method is well-suited to real-case missions as well as embedded scenarios where only a small amount of pre-training data is available and only a few labels are provided during the mission. In this situation, a consistent improvement in performance was observed when fine-tuning with visually similar data. Future work will investigate improved similarity evaluation methods as well as adaptive data-mining strategies to take advantage of the complementary pros and cons of loss-based and similarity-based mining methods.


\end{document}